%% file: main.tex
\documentclass[graybox]{svmult}

\usepackage{mathptmx}       
\usepackage{helvet}         
\usepackage{courier}        
\usepackage{type1cm}        
\usepackage{makeidx}         
\usepackage{graphicx}        
\usepackage{multicol}        
\usepackage[bottom]{footmisc}
\usepackage{hyperref}
\hypersetup{bookmarksopen,bookmarksnumbered,
pdfpagemode=UseOutlines,
colorlinks=true,
linkcolor=blue,
anchorcolor=blue,
citecolor=blue,
filecolor=blue,
menucolor=blue,
urlcolor=blue
}
\usepackage{marvosym}

\usepackage[utf8]{inputenc}
\usepackage[T1]{fontenc}    
\usepackage{upgreek}
\usepackage{amsfonts,amssymb}
\usepackage{bm}
\usepackage{bbm}

\usepackage{amsthm} 
\usepackage{thmtools}
\usepackage{mathtools}
\usepackage{epstopdf}
\usepackage{xspace}
\usepackage[numbers, sort&compress]{natbib}

\usepackage[font=footnotesize]{caption} 
\usepackage[font=footnotesize]{subcaption}
\usepackage{units}
\usepackage{booktabs} 
\usepackage{tabulary}
\usepackage[usenames,dvipsnames]{xcolor} 
\usepackage{diagbox}
\usepackage[ruled,vlined,linesnumbered]{algorithm2e}  
\usepackage{parskip}
\SetKwComment{Comment}{$\triangleright$\ }{}
\usepackage{ifthen,version}
\usepackage{soul}
\usepackage{fixltx2e}
\usepackage{csquotes}
\usepackage{microtype}      
\usepackage{lipsum}
\usepackage{csquotes}
\usepackage{tikz}

\usepackage{enumitem}

\newcommand{\xxnote}[3]{}
\ifx\hidenotes\undefined
  \usepackage{color}
  \renewcommand{\xxnote}[3]{\color{#2}{#1: #3}}
\fi

\newtheoremstyle{hypstyle}
{3pt} 
{3pt} 
{\itshape} 
{} 
{\bfseries} 
{.} 
{.5em} 
{} 

\theoremstyle{hypstyle}

\input{macros/math_operators}

\input{macros/math_defns}
\input{macros/text_shortcuts}
\graphicspath{{figs/}}

\makeindex             

\begin{document}
\title*{Monocular and Stereo Cues for Landing Zone Evaluation for Micro UAVs\vspace{-5mm}}
\vspace{-5mm}
\titlerunning{Landing Zone Evaluation for Micro UAVs}
\vspace{-10mm} \vspace{-5mm}
\author{Rohit Garg, Shichao Yang and Sebastian Scherer\\\vspace{2mm}
The Robotics Institute, Carnegie Mellon University, Pittsburgh, PA\\
email: \{rg1,shichao\}@andrew.cmu.edu, basti@cs.cmu.edu\\
}
\vspace{-5mm}
\authorrunning{R. Garg, S. Yang, and S. Scherer}
%
%
\maketitle

\vspace{-5mm}
\section{Abstract}
\vspace{-5mm}
Autonomous and safe landing is important for unmanned aerial vehicles. We present a monocular and stereo image based method for fast and accurate landing zone evaluation for UAVs in various scenarios. Many existing methods rely on Lidar or depth sensor to provide accurate and dense surface reconstruction. We utilize stereo images to evaluate the slope and monocular images to compute homography error. By combining them together, our approach works for both rigid and non-rigid dynamic surfaces. Experiments on many outdoor scenes such as water, grass and roofs, demonstrate the robustness and effectiveness of our approach.

\vspace{-5mm}
\input{inputs/1_intro.tex}

\vspace{-5mm}

\input{inputs/2_approach.tex}
\vspace{-5mm}

\input{inputs/3_experiments.tex}
\vspace{-5mm}

\input{inputs/4_insights_future.tex}
\vspace{-5mm}








  \bibliographystyle{unsrt}
  \small{\bibliography{reference}}
\end{document}

%% file: macros/math_operators.tex
\newcommand{\bbm}{\begin{bmatrix}}
\newcommand{\ebm}{\end{bmatrix}}


%% file: inputs/1_intro.tex
\section{Introduction}
\label{sec:intro}
\vspace{-5mm}

Safety is an important factor in drone autonomy. How does a drone guarantee safety for different scenarios that come up in autonomous flight? Landing is one such scenario that requires the drone to successfully evaluate potential zones that are safe to land on. 
This places a high importance on the perception system of the drone to provide accurate and robust feedback of the landing surface. 
There has been extensive work done on evaluation of unprepared landing sites for autonomous aerial vehicles. Most vision based approaches rely on a monocular/stereo camera and accurate motion estimation to compute dense surface reconstruction. Micro unmanned aerial vehicles (UAV) often operate in GPS denied environment without accurate global state estimation \cite{fang2017robust}, therefore, they are usually equipped with camera and inertial IMU sensor. Our algorithm relies only on stereo and IMU information and is designed to run fast and reliably. It is designed for the last few metres of the landing maneuver which makes it reactive to scene changes on the surface below such as people walking below. The system consists of two parts. One part analyses the surface below the drone using a custom stereo camera and inertial sensor combination. Slope and roughness are two metrics used to evaluate the terrain, and on the basis of which it is deemed safe to land on. The second part is a fast, monocular image based approach that is able to detect the presence of a non-rigid surfaces such as water or grass, which would normally be difficult for the stereo based approach to detect. We also built our own datasets consisting of 10 outdoor scenes with dense disparity ground truth to help us select an appropriate stereo matching algorithm to use in our work.

\begin{figure}[ht]
 \centering
 \includegraphics[scale=0.6]{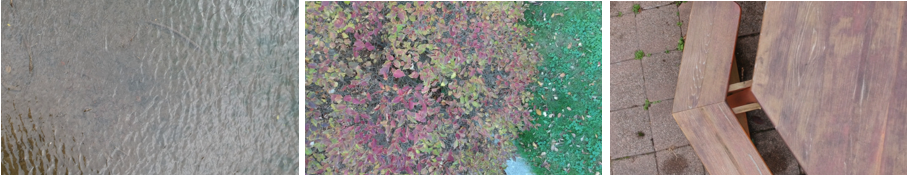}
 \caption{Various landing scenarios to evaluate including water, trees and rigid surface. Monocular and stereo cues need to be combined in order to classify them.}
 \label{fig:intro_scenarios}
\end{figure}

\vspace{-7mm}
\subsection{Related Work}
\vspace{-5mm}


There has been work done on vision based autonomous landing in unknown environments. Monocular camera based approaches can broadly be divided into two types. One type is to use a homography matrix to find feature points in a planar area \cite{bosch2006autonomous, cheng2010real}. The other approach is to build a dense point cloud using motion stereo triangulation \cite{desaraju2015vision, forster2015continuous}. These approaches can work well in static environments such as rigid ground, roof but cannot work in dynamic scenarios where epipolar geometry fails, such as flowing water, water ripples, leaves blowing on trees. To the best of the author's knowledge, there is no prior work that demonstrates a landing zone evaluation system that can detect non-rigid surfaces such as water.

There is also work done using depth sensors such as stereo cameras \cite{theodore2006flight, meingast2004vision} and LIDAR scanners \cite{ scherer2012autonomous, chamberlain2011self} to directly get a dense map for evaluating the terrain. \cite{meingast2004vision} uses a multi-frame planar parallax algorithm to produce a digital elevation map (DEM) of the terrain. The work in \cite{scherer2012autonomous} incorporates terrain/skid interaction, and the aircraft geometry in addition to terrain modeling to assess the suitability of a good landing site. \cite{maturana20153d} extends this work to an approach that uses a 3D Convolution Neural Network to assess the safety of landing zones covered in low vegetation.

%% file: inputs/2_approach.tex
\section{Approach}
\label{sec:approach}
\vspace{-5mm}

Our robot is equipped with a stereo camera that generates a dense map for general environments, however as pointed out before, there are few feature points on the water surface and stereo matching doesn't work on the dynamic surface due to water ripples caused by rotor downwash. Therefore in addition to using the stereo camera, we also use monocular cues to detect the water's surface. This section is divided into two parts on the basis of the two approaches followed.

\subsection{Monocular Approach}
\vspace{-5mm}

\begin{figure}[ht]
 \centering
 \includegraphics[scale=0.43]{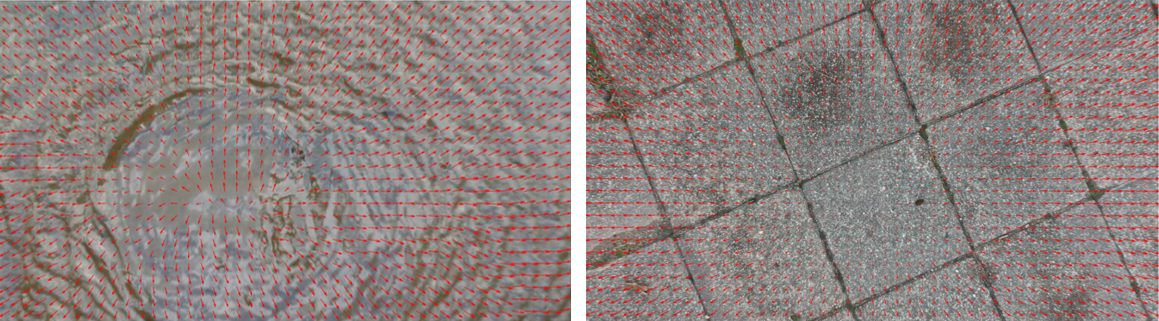}
 \includegraphics[scale=0.27]{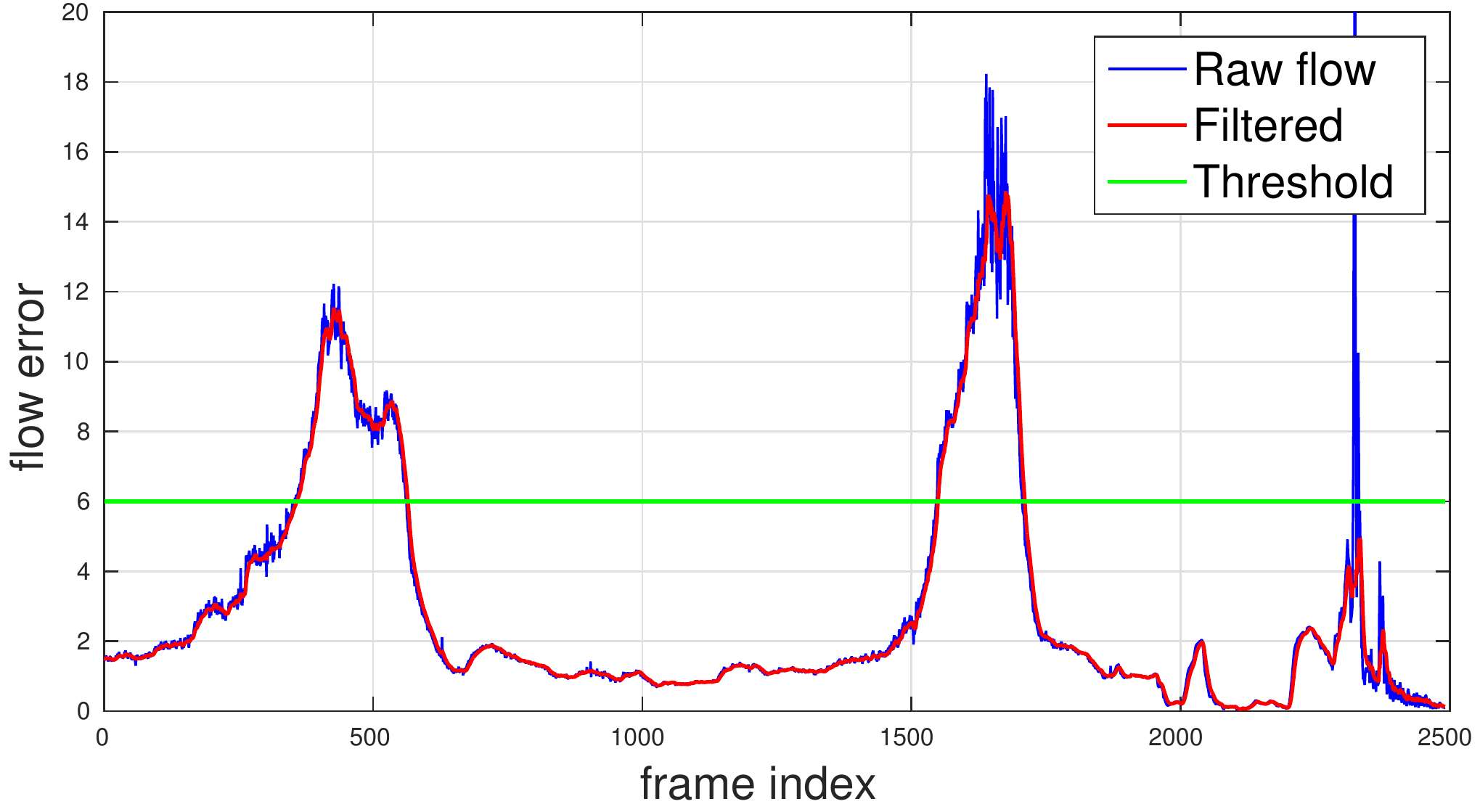} 
 \includegraphics[scale=0.27]{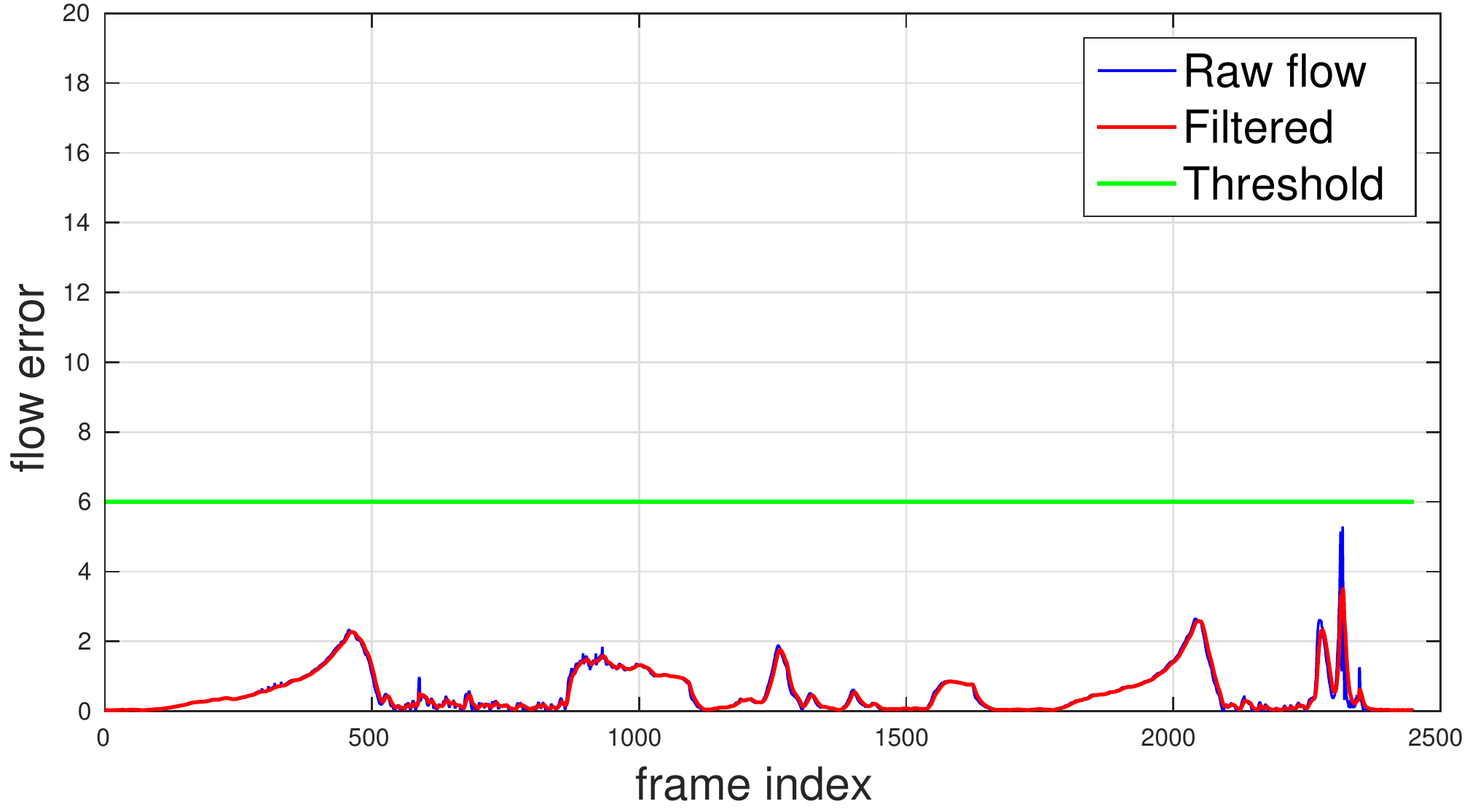}  
 \caption{ Dense optical flow pattern and planar homography error in different environments. From the bottom left, we can see that planar error increases greatly when the robot approaches the water surface and downwash from its rotors starts generating ripples. Flying over a rigid surface, the flow error doesn't change much.
 }
 \label{fig:mono flow}
\end{figure}

Similar to prior monocular work \cite{bosch2006autonomous, cheng2010real}, we use planar homography to determine the suitability of an area for safe landing. However, instead of using sparse feature matching, we choose dense optical flow to compute homography. For every two successive images, we densely sample pixels (every 20 pixels in our case) on the image denoted as $P$, then compute dense optical flow $Q$ shown in Fig \ref{fig:mono flow} using Lucas-Kanade algorithm \cite{baker2004lucas}. Finally, the planar homography can be found using all sampled pixels and the homography error $e$ indicates whether the area is planar and safe to land on. A low pass filter is applied to homography error to remove the noise. As can be seen from the bottom of Fig \ref{fig:mono flow}, there is a large homography error when the robot approaches the water surface, while on the rigid ground, the error doesn't increase much.

\begin{equation}
e = \min_H \|P-H(P+Q)\|_2
\end{equation}

\vspace{-7mm}
\subsection{Stereo Approach}
\vspace{-5mm}

\begin{figure}[ht]
 \centering
 \includegraphics[width=1\textwidth]{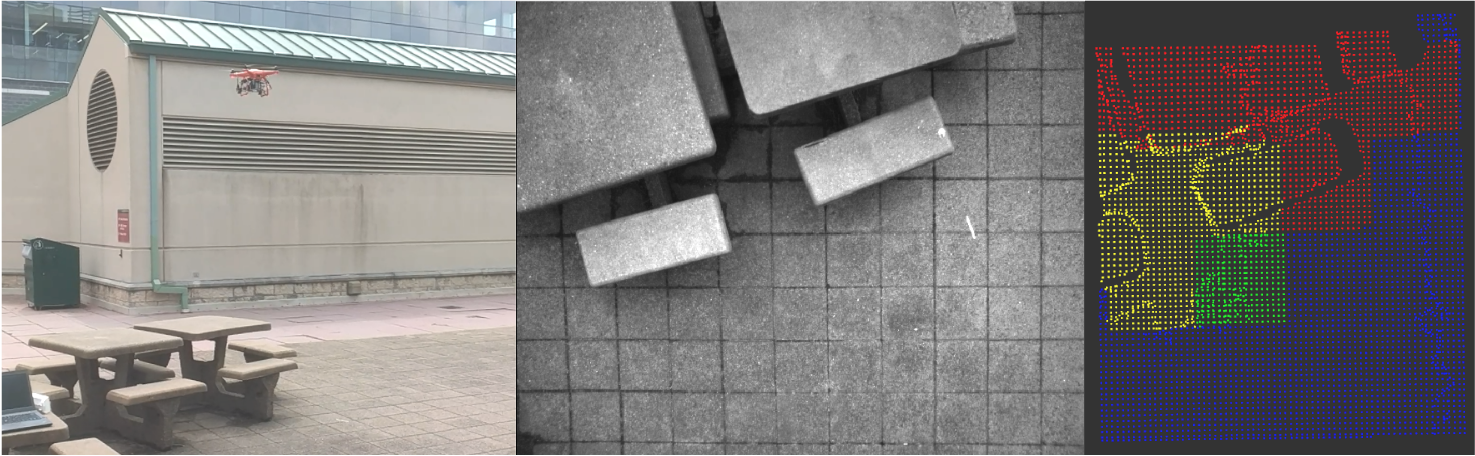}
 \caption{ Output of the stereo geometric evaluation pipeline. The drone flies over a set of tables. The middle picture shows the left camera output and the rightmost picture shows the output of the algorithm. 
 }
 \label{fig:tables_demo}
\end{figure}

This part utilizes the stereo images to compute surface depth information. This is subsequently split into a grid, and each cell within the grid is evaluated on roughness and slope metrics. (1) A fast block matching method is used to generate point clouds which are subsequently aligned to the gravitation vector using IMU information fused through a Madgwik Filter \cite{madgwick2010efficient}. (2) The point cloud is divided into a grid wherein each cell in the grid represents a 0.5 m x 0.5 m area. (3) A least squares plane fit is computed for each cell in the grid. The normal and residual are used to compute the slope and roughness of the surface within the cell. (4) A 1 sq. m. area is considered for safe landing that lies directly below the drone. Based on the roughness and slope thresholds, the algorithm filters and sends a Boolean command to the landing controller for the drone to either proceed with landing or to stop. The output from one run is shown in Fig \ref{fig:tables_demo}. The 1 sq. m. area under the drone is colored yellow for dangerous and green for safe. Outside the 1 sq. m, red indicates danger and blue indicates safety. If a part of the area under the drone is unsafe, the algorithm asks the landing controller to abort the descent.
 
%% file: inputs/3_experiments.tex
\vspace{-5mm}
\section{Experiments} 
\label{sec:experiments}
\vspace{-5mm}

This section shows the experimental results of the two proposed approaches. We also show the comparison of different stereo matching algorithms using our own stereo disparity dataset. The robot platform we have used is an Autel X-Star Quadrotor UAV. It has a custom built stereo camera setup with an IMU attached to it. The cameras and IMU are triggered together using a hardware GPIO input provided by a microcontroller.

\vspace{-7mm}
\subsection{Monocular Landing Evaluation}
\vspace{-5mm}
We evaluated the monocular approach on various water surfaces including lakes, ponds, streams, puddles and also other general environments such as rigid ground, grass, and trees. The drone flies over a target region and starts to gradually descend. It stops when the algorithm detects danger. As can be seen from Table \ref{tab:flow_results}, the algorithm is robust to variations in water surface and rigid ground surface. In scenarios with fallen leaves where we think it is safe to land, the algorithm detects large planar error due to leaves blowing away by the rotor downwash. For trees, the optical flow is scattered due to motion of the leaves. However, these are scenarios that can be handled more effectively by the Stereo based approach.

\begin{table}[ht]
\centering
 \begin{tabular}{||c | c | c | c | c||} 
 \hline
 Surface & No. of Trials & Accuracy & Robustness \\ [0.5ex] 
 \hline\hline
 Water & 15 & 100\% & Robust  \\ 
 \hline
 Solid Ground/Roof & 17 & 100\% & Robust \\
 \hline
 Grass & 4 & 100\% & Robust \\
 \hline
 Fallen Leaves & 3 & 67\% & Not Robust \\
 \hline
 Trees & 7 & 70\% & Not Robust \\ [1ex] 
 \hline
\end{tabular}
\caption{Summary of Monocular Evaluation Tests}
\label{tab:flow_results}
\end{table}

\vspace{-7mm}
\subsection{Stereo Disparity Dataset}
\vspace{-5mm}

There are existing benchmarks such as KITTI \cite{Menze2015CVPR} and Middlebury \cite{scharstein2014high} that provide an extensive set of scenes for evaluation of stereo algorithms. However, they don't cover cases with a short baseline stereo and downward looking scenes. This motivated us to build our own dataset consisting of 10 scenes taken in different locations around the CMU campus. The dense ground truth disparity map is computed from data acquired from accurate Faro 120 Laser Scanner. Stereo images were captured at 2208 x 1242 resolution using the ZED Stereo Camera. The results are summarized from comparing 5 algorithms: Block Matching (BM)\cite{konolige1998small}, BM + WLS Filter + LR Check, Semi-Global Matching (SGBM) \cite{hirschmuller2005accurate}, SGBM + WLS Filter + LR Check, and LIBELAS\cite{Geiger2010ACCV} shown in Table \ref{tab:bad_4}. The results are mixed in terms of accuracy. SGBM + WLS + LRC is a close match for LIBELAS. But both methods are quite slow to run the NVIDIA TX2 platform. We finally chose Block Matching because it is the fastest out of all 5 and would ensure the entire pipeline could be run in real-time.


\begin{table}[ht]
\centering
 \begin{tabular}{||c | c | c | c | c | c |} 
 \hline
 Scene & BM & BM + WLS + LRC & SGBM & SGBM + WLS + LRC & LIBELAS \\ [0.5ex] 
 \hline\hline
 Scene 1 & 33.3\% & 30.3\% & 18\% & 16.2\% & 9.8\%  \\ 
 \hline
 Scene 2 & 37.6\% & 36.5\% & 21.8\% & 20.9\% & 20.2\%  \\
 \hline
 Scene 3 & 37.4\% & 36.2\% & 18.8\% & 17.6\% & 12.6\%  \\
 \hline
 Scene 4 & 39.1\% & 39.5\% & 17.6\% & 16.6\% & 9.1\%  \\
 \hline
 Scene 5 & 42.5\% & 37.2\% & 17.8\% & 16.0\% & 23.7\%  \\
 \hline
 Scene 6 & 67.0\% & 53.2\% & 33.1\% & 31.5\% & 40.4\%  \\
 \hline
 Scene 7 & 70.0\% & 58.3\% & 30.2\% & 28.8\% & 40.5\%  \\
 \hline
 Scene 8 & 49.2\% & 45.5\% & 35.3\% & 34.4\% & 32.4\%  \\
 \hline
 Scene 9 & 54.0\% & 50.5\% & 35.8\% & 35.3\% & 29.4\%  \\
 \hline
 Scene 10 & 57.3\% & 48.7\% & 38.7\% & 36.0\% & 31.7\%  \\ [1ex] 
 \hline
\end{tabular}
\caption{Percentage of bad pixels (disparity error $> 4$ pixel)}
\label{tab:bad_4}
\end{table}

\vspace{-7mm}
\subsection{Stereo Landing Evaluation}
\vspace{-5mm} 
Similar to monocular landing evaluation, we perform landing tests in different scenes using the stereo based approach. The robot will stop descending when the stereo module detects danger. Results are shown in Table \ref{tab:geom_results}.


Our approach can work well in most environments except in areas with textureless surfaces where stereo matching is not able to find enough matches to do a good reconstruction. Tall, sharp, and thin grass is also difficult as it only occupies small regions in the image making it difficult for the stereo matching to find enough correspondences.


\begin{table}[ht]
\centering
 \begin{tabular}{||c | c | c | c |} 
 \hline
 Surface/Obstacle & No. of Trials & Success Rate & Description \\ [0.5ex] 
 \hline\hline
 Short Grass & 5 & 100\% & Grass blades not longer than 10 cm  \\ 
 \hline
 Tarmac & 5 & 100\% & Metalled Road surface  \\
 \hline
 Chair & 5 & 100\% & Outdoor lounge chairs placed in a grassy area \\
 \hline
 Box & 5 & 100\% & Plastic storage box (dimensions 0.3x0.4x0.3m)  \\
 \hline
 Tables & 5 & 100\% & Outdoor concrete tables with benches  \\
 \hline
 Concrete & 5 & 100\% & Concrete flat ground \\
 \hline 
 Steps & 5 & 100\% & Outdoor concrete staircase \\
 \hline
 Stones & 5 & 100\% & Large stones (Elliptical shape - major axis 0.3 m) \\
 \hline
 Tall Grass & 5 & 0\% & Tall grass blades longer than 15 cm \\
 \hline
 Textureless Surfaces & 5 & 0\% & Uniformly colored Volleyball Ground\\ [1ex] 
 \hline
\end{tabular}
\caption{Summary of Geometric Evaluation Tests}
\label{tab:geom_results}
\end{table}

%% file: inputs/4_insights_future.tex
\vspace{-5mm}
\section{Conclusions} 
\label{sec:insights}
\vspace{-5mm}

In this paper, we propose a vision based landing evaluation using monocular and stereo cues. Stereo enables a dense surface reconstruction in most environments to evaluate slope and roughness, however it cannot work well in challenging low-texture scenes as well as non-rigid surfaces. A monocular approach that exploits optical flow in the scene is proposed to handle such cases. Results from the experiments show that the two approaches are fairly robust to different scenes. Their respective failure scenarios also indicate that the two approaches can complement each another.

Currently we have tested the two approaches separately. We will combine them into a single module to assess the landing zone automatically and robustly.

